\documentclass[runningheads]{llncs}

\usepackage[T1]{fontenc}
\usepackage{graphicx}
\usepackage{hyperref}
\usepackage{color}

\usepackage{listings}

\usepackage{makecell}

\usepackage{tabularx}
\usepackage[dvipsnames]{xcolor}
\usepackage{multirow}

\usepackage{array}
\usepackage{makecell}

\usepackage{array}
\newcolumntype{P}[1]{>{\centering\arraybackslash}p{#1}}

\begin{document}
\title{Studying the role of named entities for \\ content preservation in text style transfer  
}

\author{Nikolay Babakov$^1$ \and David Dale$^1$ \and  Varvara Logacheva$^1$ \and Irina Krotova$^2$ \and 
Alexander Panchenko$^1$}

\authorrunning{Babakov et al.}

\titlerunning{Role of named entities for content preservation in text style transfer}

%
\institute{Skolkovo Institute of Science and Technology
\email{\{n.babakov,d.dale,a.panchenko\}@skoltech.ru}
\and
Mobile TeleSystems (MTS)
}

%
\maketitle              
\begin{abstract}

Text style transfer techniques are gaining popularity in Natural Language Processing, finding various applications such as text detoxification, sentiment, or formality transfer. However, the majority of the existing approaches were tested on such domains as online communications on public platforms, music, or entertainment yet none of them were applied to the domains which are typical for task-oriented production systems, such as personal plans arrangements (e.g. booking of flights or reserving a table in a restaurant). We fill this gap by studying formality transfer in this domain. 

We noted that, the texts in this domain are full of named entities, which are very important for keeping the original sense of the text. Indeed, if for example, someone communicates destination city of a flight is must not be altered. Thus, we concentrate on the role of named entities in content preservation for formality text style transfer.

We collect a new dataset for the evaluation of content similarity measures in text style transfer. It is taken from a corpus of task-oriented dialogues and contains many important entities related to realistic requests that make this dataset particularly useful for testing style transfer models before using them in production. Besides, we perform an error analysis of a pre-trained formality transfer model and introduce a simple technique to use information about named entities to enhance the performance of baseline content similarity measures used in text style transfer.





\end{abstract}

\section{Introduction}

Text style transfer (\textbf{TST}) systems are designed to change the style of the original text to alternative one, such as more informal~\cite{rao-tetreault-2018-dear}, more positive~\cite{luo-etal-2019-towards}, or even more Shakespearean~\cite{jhamtani-etal-2017-shakespearizing}. Such systems have gained significant popularity in the NLP within the last few years. They could be applied to many purposes: from diversifying responses of dialogue agents to creating artificial personalities.

More formally, TST system is a function $\alpha:~S~\times~S~\times~D~\rightarrow~D$ that, given a source style $s^{src}$, a target style $s^{dst}$, and an input text $d^{src}$, produces an output text $d^{dst}$ \mbox{such that:}
\begin{itemize}

\item The style of the text changes from the source style $s^{src}$ to the target style $s^{dst}: \sigma(d^{src}) \neq \sigma(d^{dst})$, $\sigma(d^{dst}) = s^{dst}$;

\item The content of the source text is saved in the target text as much as required for the task: $\delta(d^{src}, d^{dst}) \geq t^{\delta}$;
\item The fluency of the target text achieves the required level: $\psi(d^{dst}) \geq t^{\psi}$,

\end{itemize}


where $t^{\delta}$ and $t^{\psi}$ are task-specific thresholds for the content preservation ($\delta$) and fluency ($\psi$) functions. 


To measure if the content of the source text $d^{src}$ is preserved in the target text $d^{dst}$ a \textbf{content similarity measure} is used.  This is a specific similarity measure $sim$ which quantifies semantic
relatedness of $d^{src}$ and $d^{dst}$ : $sim(d^{src}, d^{dst})$. The measure $sim$ yields high score for the pairs with similar content and low score for ones with different content.




In the majority of recent TST papers~\cite{DBLP:conf/sigir/RaneDLE21,cao-etal-2020-expertise,riley-etal-2021-textsettr}
BLEU~\cite{papineni2002bleu} is still the main way to evaluate the content similarity. More recent approaches as cosine similarity calculation between averaged word vectors~\cite{pang-gimpel-2019-unsupervised}, 
BLEURT~\cite{sellam-etal-2020-bleurt} (which is a BERT~\cite{devlin-etal-2019-bert} fine-tuned for semantic similarity evaluation task in cross-encoder manner on synthetic data)~\cite{lai-etal-2021-thank} and BERTScore\cite{zhang2019bertscore} (F1-score over BERT-embeddings between tokens from initial and target sentences)~\cite{lee-etal-2021-enhancing} are also gaining popularity. 


To the best of our knowledge, none of the newly proposed TST techniques have been tested in the domain of the personal plan. We consider the step towards such a domain in TST research valuable because it makes its application in the real world even more likely. One of the main distinguishing properties of this domain is a large number of named entities (\textbf{NE}). NEs are real-world objects, such as a person, location, organization, etc. Indeed, when a client wants to order a taxi or book a flight, and a dialogue agent's reply is modified to, for example, a more informal style to make a conversation more natural, it is crucial to keep all significant details of the client's request, as a destination of a taxi ride or a name of the departure airport.

We assume that if a NE is lost during TST, then some important parts of the original content are lost. For example, in~\cite{nema-khapra-2018-towards} authors exploited a similar assumption and used the information about NEs and some other categories of words to improve measures like BLEU or METEOR~\cite{banerjee-lavie-2005-meteor} for question answering task. Thus, we dedicate our work to studying the role of named entities and other linguistic objects in the process of TST and, in particular, in content similarity scoring.

The contributions of our paper are as follows:
\begin{itemize}
    \item We create and release\footnote{\url{https://github.com/skoltech-nlp/SGDD-TST}} the first benchmark dataset for evaluating content similarity measures in style transfer in the task-oriented dialogue domain (Section~\ref{section:dataset});
    
    \item We perform an error analysis of a SOTA pre-trained text style transfer system in terms of content preservation (Section~\ref{sec:info_loss});
    
    \item We perform an error analysis of SOTA content similarity measures used in text style transfer (Section~\ref{sec:measures_errors});
    
    \item We introduce a simple technique for enriching the content similarity measures with information about named entities, which increases the quality of strong baseline measures used in text style transfer (Section~\ref{sec:ner}). 
\end{itemize}

\section{Dataset collection}
\label{section:dataset}


In this section, we describe the process of collection of SGDD-TST (Schema-Guided Dialogue Dataset for Text Style Transfer) -- a dataset for evaluating the quality of content similarity measures for text style transfer in the domain of the personal plans. We use a pre-trained formality transfer model to create style transfer paraphrases from the task-oriented dialogue dataset. The obtained sentence pairs are then annotated with similarity labels by crowd workers.


\subsection{Generation of parallel texts}



One of the core contributions of our work is the new TST dataset annotated with content similarity labels. The topics of most of the existing TST formality datasets~\cite{briakou-etal-2021-evaluating,krishna-etal-2020-reformulating,Yamshchikov_Shibaev_Khlebnikov_Tikhonov_2021} are mostly related to common discussion of entertainment or family affairs~\cite{rao-tetreault-2018-dear}, so it could be more useful to calculate content similarity for TST applied to real-world tasks, such as booking hotels or purchasing tickets. 

The initial data for the dataset was collected from SGDD (Schema-Guided Dialogue Dataset)~\cite{rastogi2020towards}. This dataset consists of 16,142 task-oriented dialogues, which are naturally full of NEs related to real-life tasks (booking of hotels, flights, restaurants). If a NE is lost or corrupted during TST the overall sense of the initial sentence is most probably lost as well.
This makes the style transferred pairs particularly interesting for the task of content preservation control.

\begin{figure}[t!]
\centering
\includegraphics[scale=0.6]{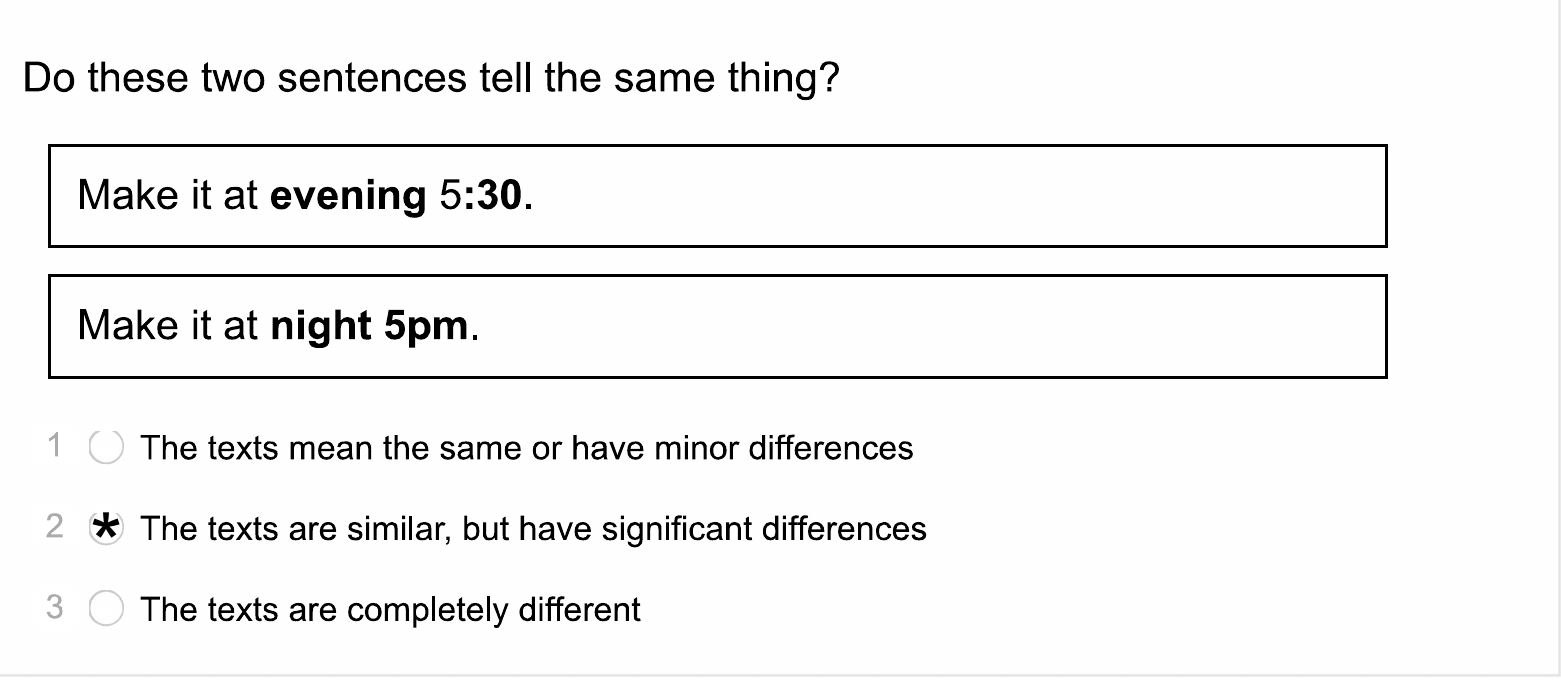}
\caption{The interface of the content similarity crowdsourcing task.}
\label{fig:yandex_interface}
\end{figure}

As far as SGDD is not originally related to TST, we use a base T5 model \cite{t5}\footnote{\url{https://huggingface.co/ceshine/t5-paraphrase-paws-msrp-opinosis}} fine-tuned for 3 epochs with learning rate $1e^{-5}$ with Adam optimizer on parallel GYAFC formality dataset~\cite{rao-tetreault-2018-dear} to generate style transferred paraphrases. 




\subsection{Annotation setup}
\label{sub:annot_setup}

The crowdsourcing report is performed according to recommendations by~\cite{briakou-etal-2021-review}, where the authors propose a standardized way to open-source the details of collected TST datasets.


We need to choose the scale to evaluate the semantic similarity. For example in~\cite{Yamshchikov_Shibaev_Khlebnikov_Tikhonov_2021} 5 labels were used, but the final agreement by Krippendorf's alpha~\cite{krippendorf} coefficient is rather low: 0.34. Thus we use only three labels. An example of a crowdsourcing task interface can be found in Figure~\ref{fig:yandex_interface}.



The annotation is performed with the Yandex.Toloka\footnote{\url{https://toloka.yandex.com}} platform. To prepare the workers we use training tasks (with pre-defined answers and explanations) and control tasks (with pre-defined answers and without explanations). Workers are admitted to the real tasks after solving training and control tasks with acceptable grades. These tasks are also merged with the real ones. If the workers fail to pass them, they are banned and all their annotations are discarded.

\subsection{Dataset statistics}

\begin{figure}[t]
\centering
\includegraphics[scale=0.45]{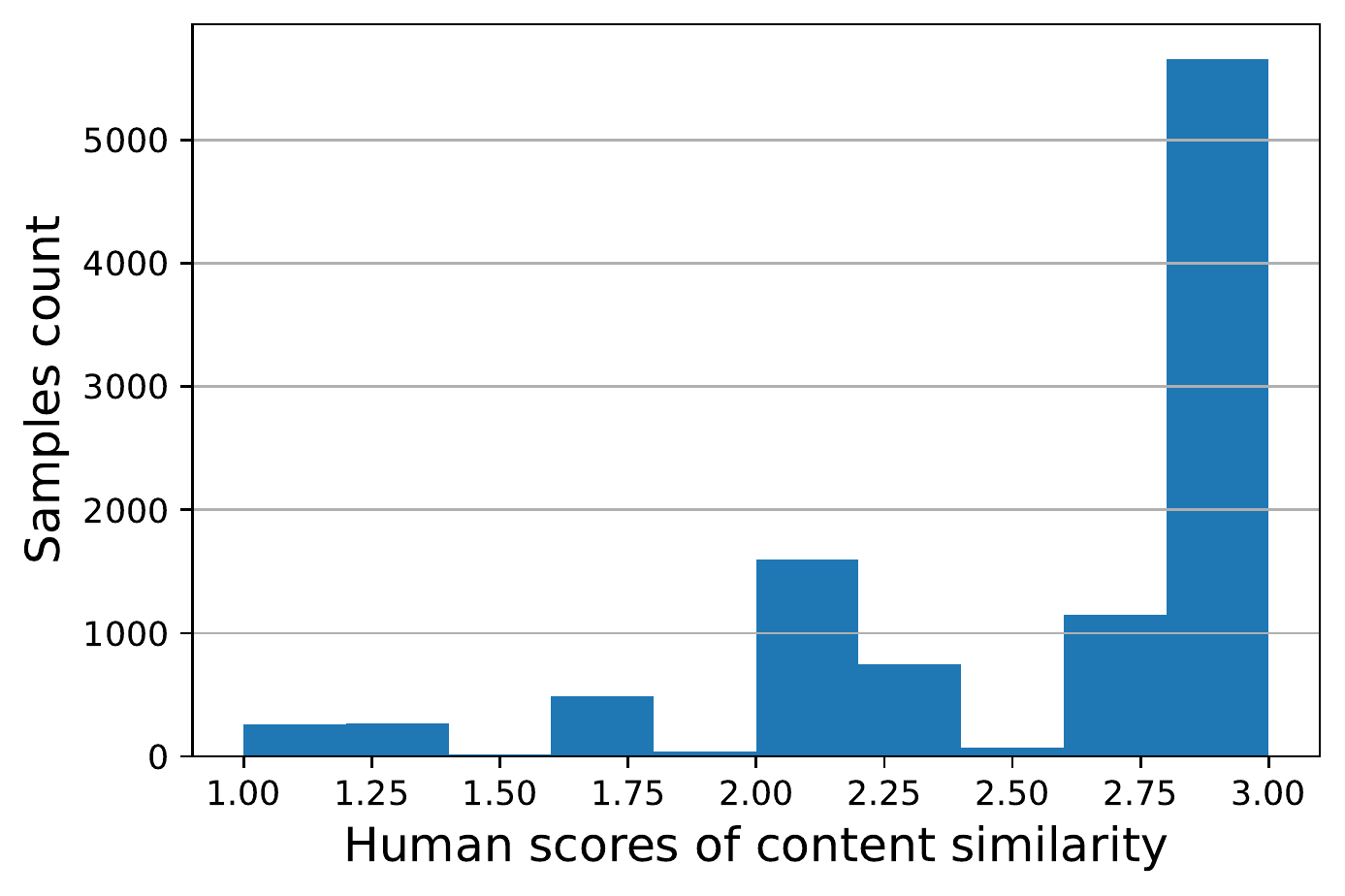}
\caption{Distribution of averaged human similarity score between original phrase and the phrase generated with the TST model. In most cases, the content in the text rewritten with the model is not lost.}
\label{fig:score_distribution}
\end{figure}

The final size of the dataset is 10,287 text pairs. The final similarity score for each pair was obtained by simply averaging the votes, where 1 point stands for ``The texts are completely different'', 2 stands for ``The texts are similar but have significant differences'', and 3 stands for ``The texts mean the same or have minor differences''. The distribution of the scores in the collected dataset can be found in Figure~\ref{fig:score_distribution}. Some samples from the dataset are shown in Table~\ref{tab:dataset_samples}.

\begin{table}[t]
\centering
\resizebox{\textwidth}{!}{
\begin{tabular}{|P{0.35\linewidth}|P{0.35\linewidth}|P{0.15\linewidth}|P{0.15\linewidth}|P{0.12\linewidth}|P{0.05\linewidth}|}
\hline
\textbf{Original text} & \textbf{Generated text} & \textbf{\#Different} & \textbf{\#Similar} & \textbf{\#Same} &
\textbf{Sim} \\ 
\hline
Where are you planning to leave from? & Where are you from and where is your plane? & 3 & 0 & 0 & 1 \\ \hline
I will depart from Vancouver. & i'll go to Vancouver and get out of there. & 2 & 0 & 1 & 1.66 \\ \hline
I have found 3 restaurants for you, one of which is called Locanda Positano which is located in Lafayette. & I have found three restaurants for you, one is called Locanda Positano. & 0 & 3 & 0 & 2 \\ \hline
I need a roundtrip flight departing from LAX please on any airline. & I need a roundtrip flight departing from Los Angeles, please. & 0 & 1 & 2 & 2.66 \\ \hline
Yes, I'd like to book the tickets. & yes i wanna book the tickets & 0 & 0 & 3 & 3 \\ 
\hline
\end{tabular}}
\caption{Samples from the collected SGDD-TST dataset. Columns from third to fifth indicate the number of votes for one of the answers to the crowdsourcing task: \textbf{\#Different} - ``The texts are completely different'', \textbf{\#Similar} - ``The texts are similar, but have significant difference'', and \textbf{\#Same} -``The texts mean the same or have minor difference''. \textbf{Sim} shows the averaged human scores.}
\label{tab:dataset_samples}
\end{table}

The dataset was annotated by 1,214 workers (3 to 6 workers per sample). Krippendorf's alpha agreement score is 0.64. The average task annotation time is 15.7 seconds. The average percentage of right answers to control and training tasks merged with real tasks is 0.65.   


There are similar datasets with human annotation about content similarity collected for different TST tasks: detoxification (Tox600~\cite{dementieva2021crowdsourcing}), sentiment transfer (Yam.Yelp~\cite{Yamshchikov_Shibaev_Khlebnikov_Tikhonov_2021}), and formality transfer (xformal-FoST~\cite{briakou-etal-2021-evaluating}, STRAP~\cite{krishna-etal-2020-reformulating}, and Yam.GYAFC~\cite{Yamshchikov_Shibaev_Khlebnikov_Tikhonov_2021}).
To the best of our knowledge, our dataset is the biggest TST dataset with human annotations of content similarity (see Table~\ref{tab:comparison_other_datasets}). 
Moreover, while the existing formality transfer datasets are based on GYAFC~\cite{rao-tetreault-2018-dear} collected by formal rewrites of phrases from Yahoo Answers L6 corpus,~\footnote{\url{https://webscope.sandbox.yahoo.com/?guccounter=1}} our dataset is the first one based on task-oriented dialogues, which allows making another step towards applying TST, in particular formality transfer, to real-world tasks.


\begin{table}[t!]
\centering
\resizebox{\textwidth}{!}{

\begin{tabular}{|c|cccc|c|c|}
\hline
\textbf{Name} & \multicolumn{1}{c|}{xformal-FoST} & \multicolumn{1}{c|}{Yam. GYAFC} & \multicolumn{1}{c|}{STRAP. GYAFC} & Tox600 & Yam. Yelp & SGDD-TST \\ \hline
\textbf{Size} & \multicolumn{1}{c|}{2,458} & \multicolumn{1}{c|}{6,000} & \multicolumn{1}{c|}{684} & 600 & 2,000 & 10,287 \\ \hline
\textbf{Task} & \multicolumn{3}{c|}{Formality transfer} & Detoxification & Sentiment transfer & \makecell{Task oriented \\ formality transfer} \\ \hline
\textbf{Domain} & \multicolumn{4}{c|}{Online communication about recreational topics} & Service review & Personal plans \\ \hline
\end{tabular}

}
\caption{Comparison of our SGDD-TST dataset with other TST datasets.}
\label{tab:comparison_other_datasets}
\end{table}

\section{Error analysis of the pre-trained text style transfer system}
\label{sec:info_loss}

In this section, we try to understand what kind of errors occur when a large TST model pre-trained on parallel data generates a new utterance that is considered different from the initial one by most human annotators.

\subsection{Experimental setting}
\label{sec:info_loss_setting}

We annotate 400 random pairs from the collected dataset which are annotated as not semantically equal by crowd workers. We use the following categories for annotation, which in most cases are mutually exclusive :
\begin{itemize}
    \item \textbf{Named entities} We check whether the loss or corruption of NEs yields the content loss. In SGDD NEs are mostly related to time, places, and other objects used in different kinds of services;
    \item \textbf{Lost parts of speech} We check whether loss or corruption of some specific parts of speech (POS) not related to NEs affect the content of generated sentence;
    \item \textbf{Corrupted sentence type} We check whether the original sentence is related to one sentence type (declarative, imperative, interrogative, and exclamatory) and the newly generated one becomes related to another one. 
\end{itemize}
 



\subsection{Results}

\begin{figure}[]
\centering
\includegraphics[scale=0.4]{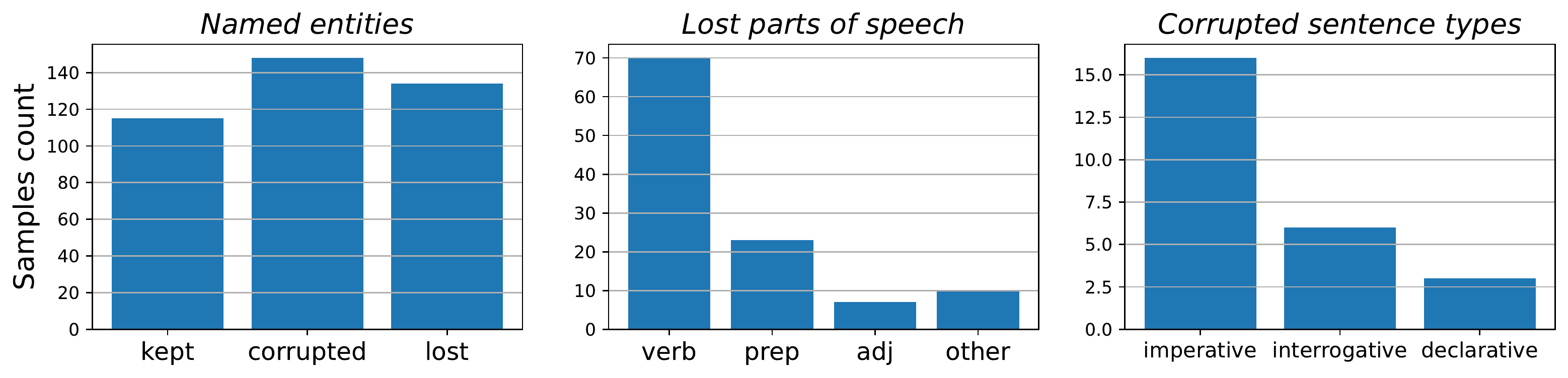}
\caption{Statistics of different reasons of content loss in TST.}
\label{fig:phenomen_barcharts}
\end{figure}

\begin{figure}[]
\centering
\includegraphics[scale=0.45]{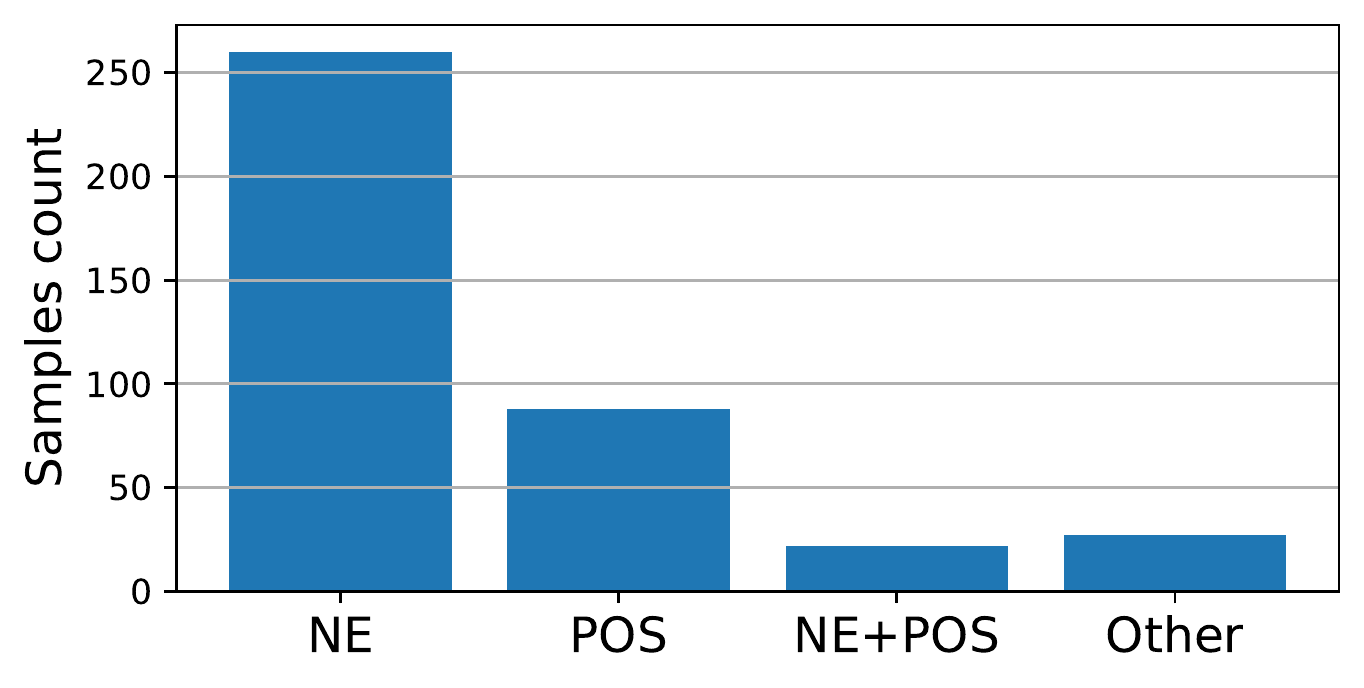}
\caption{Frequency of the reasons for the change of content between original and generated sentences: named entities (NE), parts of speech (POS), named entities with parts of speech (NE+POS), and other reasons (Other).}
\label{fig:phenomen_proportion}
\end{figure}

We show main results of the analysis in Figures~\ref{fig:phenomen_barcharts} and~\ref{fig:phenomen_proportion}. Loss or corruption of NEs is present in the vast majority of cases when the content is lost. Moreover, the loss of significant verbs or prepositions or skipping several words which change the type of sentence (e.g. from imperative to declarative) could also change the final sense of the original utterance and spoil a client's experience with a dialogue agent. See Table~\ref{tab:failures_examples} for examples.

\begin{table}[t!]
\resizebox{\textwidth}{!}{
\begin{tabular}{|P{0.4\linewidth}|P{0.4\linewidth}|P{0.18\linewidth}|P{0.15\linewidth}|P{0.15\linewidth}|}
\hline
\textbf{Original text} & \textbf{Generated text} & \textbf{Lost or corrupted NEs} & \textbf{Lost or corrupted POS}  & \textbf{Corrupted sentence type}\\ \hline
I will arrive next Thursday and \textbf{depart} on the 14th of March. & I will arrive on the 14th of March next Thursday. & Kept & Verb & Not corrupted  \\ \hline
In Paris on the 1st \textbf{until} Saturday this week. & In Paris on the first Saturday of this week. & Kept & Preposition & Not corrupted \\ \hline
I am looking for a \textbf{unisex} salon in SFO. & I am looking for a \textbf{non-isex }salon in San Francisco. & Kept & Adjective  &  Not corrupted \\ \hline
Hello, I need a bus to Sacramento \textbf{from Fresno }on the 5th of March. & Hello, I need a bus to Sacramento on the 5th of March. & Lost & Kept & Not corrupted \\ \hline
Move one thousand \textbf{two hundred} and forty dollars. & move one thousand and forty bucks & Corrupted & Kept & Not corrupted  \\ \hline
\textbf{Can you please confirm that} you need 3 rooms for the reservation on March 1st? & you need 3 rooms for the reservation on march 1st. & Kept & Kept & Interrogative \\ \hline
\end{tabular}}
\caption{Examples of different reasons for changed content in the generated text.}
\label{tab:failures_examples}
\end{table}

\section{Error analysis of content similarity measures}
\label{sec:measures_errors}

In this section, we analyze the failures of SOTA text similarity measures.

\subsection{Content similarity measures used in our study}
\label{sec:measure_definite_into}

We use both commonly used and recently proposed SOTA content similarity measures based on different calculation logic:



\begin{itemize}
    \item Word and character ngrams-based (\textbf{ngram}): BLEU, METEOR~\cite{banerjee-lavie-2005-meteor}, ROUGE based on unigrams/bigrams/trigrams/longest common sequence  (ROUGE-1/2/3/L)~\cite{lin2004rouge}, chrf~\cite{popovic-2015-chrf};
    \item Averaged word vectors similarity (\textbf{vect-sim}): Word2vec~\cite{mikolov2017advances},  Fasttext~\cite{bojanowski-etal-2017-enriching};
    \item Large pre-trained models (\textbf{pre-trained}): BERTScore (modifications based on DeBERTa~\cite{he2021deberta}, RoBERTa~\cite{zhuang-etal-2021-robustly} and BERT~\cite{devlin-etal-2019-bert}),  BLEURT\footnote{\url{https://huggingface.co/Elron/bleurt-large-512}}.
\end{itemize}

\subsection{Experimental setting}
\label{sec:measures_errors_setting}
 
We produce two rankings of sentences : a ranking based on their automatic scores and another one based on the manual scores, then sort the sentences by the absolute difference between their automatic and manual ranks, so the sentences scored worse with automatic measures are at the top of the list. We manually annotate the top 35 samples for a subset of the measures described in Section~\ref{sec:measure_definite_into} based on various calculation logic. The annotation setup is similar to the Section~\ref{sec:info_loss_setting}.



\subsection{Results}

\begin{figure}[]
\centering
\includegraphics[scale=0.38]{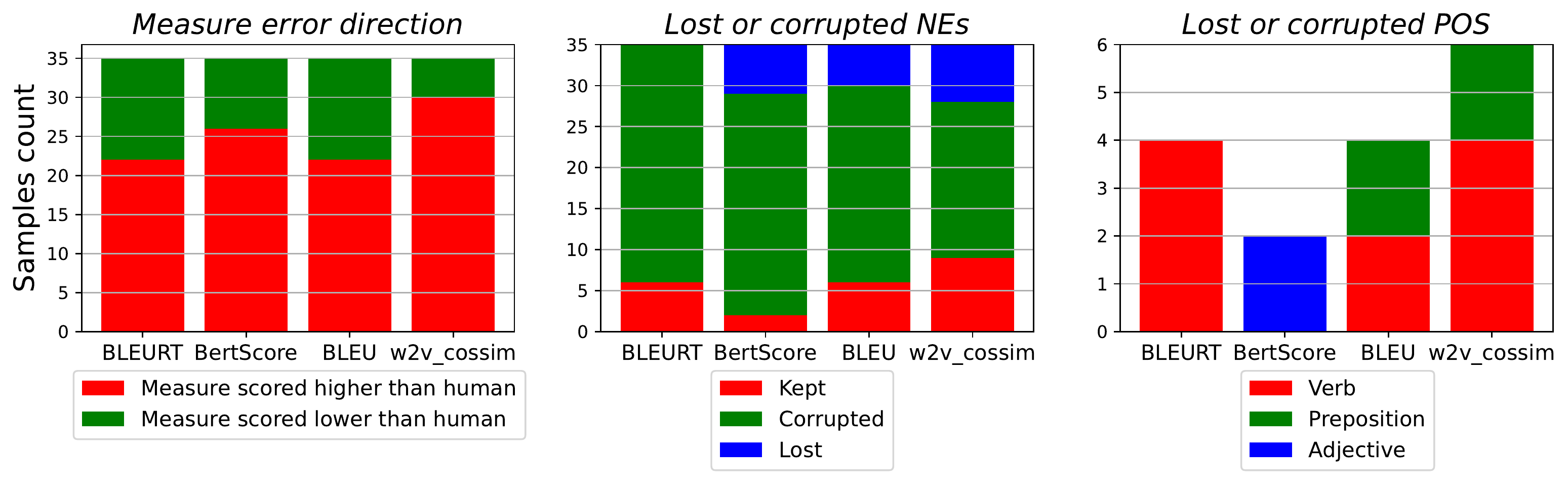}
\caption{Errors statistics of the analyzed measures. BertScore/DeBERTa is referred as BertScore here.}
\label{fig:info_loss_per_measure}
\end{figure}

We plot the results of manual annotation in Figure~\ref{fig:info_loss_per_measure}. The loss or corruption of NEs take a significant part in the failure of all measures. The loss of POS also takes part in the content loss, however, it looks much less significant. We don't report sentence type change here because this kind of change is almost absent in the analyzed samples. We also report the statistics of whether the measures scored a definite pair higher or lower than humans (to check this we apply a linear transformation to both automatic measures and human judgments so that their values are distributed between 0 and 1). In most cases of poor performance, measures assign higher scores than human annotators.

The examples of different types of content loss are shown in Table~\ref{tab:failures_examples}. In the most cases of worst measures performance, the original and generated sentences look almost the same but loss or corruption of NEs or POS play a significant role in the general meaning of the original sentence, which can be properly captured by human annotators and is hard to be captured with automatic measures.  

\section{Named entities based content similarity measure}
\label{sec:ner}

In this section, we show how to use NEs for improving SOTA similarity measures.

\subsection{Baseline named entities based approach}


Our findings in Sections~\ref{sec:info_loss} and~\ref{sec:measures_errors} show that NEs play a significant role in the content loss, thus we try to improve existing measures with NE-based signals. To make the results of this analysis more generalizable we use the simple open-sourced \href{https://github.com/explosion/spacy-models/releases/tag/en_core_web_md-3.2.0}{Spacy NER-tagger} to extract entities from the collected dataset. These entities are processed with lemmatization and then used to calculate the Jaccard index~\cite{jaccard_index} over the intersection between entities from original and generated sentences. This score is used as a baseline NE-based content similarity measure. We use Spearman Rank Correlation Coefficient between human judgments and automatic scores to check the quality of content similarity measures.




\subsection{Merging named entities based measure with other measures}

The baseline NE-based measure has a low Spearman correlation with human scores – 0.06, so we use it as an auxiliary signal by merging two signals using the following formula: $M_{weigted} = M_{strong}\times (1-p) + M_{NE}\times p$ where $p$ is a percentage of NE tokens within all tokens in both texts, $M_{strong}$ is an initial measure and $M_{NE}$ is a NE-based signal. The intuition behind the formula is that the NE-based auxiliary signal is useful in the proportion equal to the proportion of NEs tokens in the text. Thus, the score of the main measure is not changed if there are no NEs in the text, and at the same time, the more NEs are in the text the more significantly the NE-based signal will affect the main score.  We apply such merging to all measures presented in Section~\ref{sec:measure_definite_into}.

\begin{table}[t!]
\centering
\resizebox{\textwidth}{!}{
\begin{tabular}{|P{0.3\linewidth}|P{0.25\linewidth}|P{0.2\linewidth}|P{0.15\linewidth}|P{0.2\linewidth}|}
\hline
\textbf{Similarity measure} & \textbf{Measure type} & \textbf{without NE} & \textbf{with NE} & \textbf{Improvement} \\
\hline
BLEURT       &  pre-trained &      0.56 &    0.56 &      0.00 \\
BertScore/DeBERTa &  pre-trained &      0.47 &    0.45 &    \textcolor{red}{--0.02} \\
BertScore/RoBERTa &  pre-trained &      0.39 &    0.37 &    \textcolor{red}{--0.02} \\
BLEU              &        ngram &      0.35 &    0.38 &  \textcolor{OliveGreen}{+0.03} \\
ROUGE-1           &        ngram &      0.29 &    0.36 &  \textcolor{OliveGreen}{+0.07} \\
BertScore/BERT    &  pre-trained &      0.28 &    0.36 &  \textcolor{OliveGreen}{+0.08} \\
ROUGE-L           &        ngram &      0.27 &    0.35 &  \textcolor{OliveGreen}{+0.08} \\
chrf              &        ngram &      0.27 &     0.30 &  \textcolor{OliveGreen}{+0.03} \\
w2v\_cossim        &     vect-sim &      0.22 &    0.33 &  \textcolor{OliveGreen}{+0.11} \\
fasttext\_cossim   &     vect-sim &      0.22 &    0.32 &   \textcolor{OliveGreen}{+0.10} \\
ROUGE-2           &        ngram &      0.15 &    0.22 &  \textcolor{OliveGreen}{+0.07} \\
METEOR            &        ngram &       0.10 &    0.25 &  \textcolor{OliveGreen}{+0.15} \\
ROUGE-3           &        ngram &      0.09 &    0.14 &  \textcolor{OliveGreen}{+0.05} \\
\hline
\end{tabular}}
\caption{Spearman correlation of automatic content similarity measures with human content similarity scores with and without using auxiliary NE-based measure on the collected SGDD-TST dataset. }
\label{tab:ner_experiments}
\end{table}

\subsection{Results}



The results of the proposed approach are in Table~\ref{tab:ner_experiments}. All baseline measures (BLEU, ROUGE, METEOR) and some recent approaches (e.g. similarity between averaged embeddings) gain significant improvement from using this kind of auxiliary signal (the significance is measured with Williams test~\cite{graham-baldwin-2014-testing}). The most probable reason for this is that neither ngram-based measures nor vectors similarity-based measures process the information about the specific role of the NEs. However, the most modern trained measures like BLEURT and BertScore do not get any improvement from this approach. These approaches are based on large pre-trained models, so it is very likely that during training the models learned the concept of NEs and additional information can be not useful or in some cases even decrease the performance.  Even though the research performed in Section~\ref{sec:measures_errors} shows that the failures of top-performing measures are mostly related to loss or corruption of NEs, it seems that such a straightforward approach to enriching these measures with NE-related signal is not effective.




\section{Conclusion}

In this work, we collect the dataset for content similarity evaluation in text style transfer for the task-oriented dialogue domain. During the manual analysis of the collected dataset, we show that named entities play important role in the problem of content loss during text style transfer. 

We show that such baseline content similarity measures as BLEU, METEOR, and ROUGE and even more recent approaches like cosine similarity between word2vec or fasttext embeddings fail to track perturbations of such entities, thus enriching these measures with named entities-based signal significantly improves their correlation with human judgments. At the same time, with the most recent approaches such as BLEURT or BERTScore, this kind of enrichment does not yield any improvement, and the correlations of these measures with human judgments are much higher than that of baseline approaches.

However, even the top-performing measures are still far from perfect in terms of the absolute value of correlation with human labels. Thus, the collected dataset with annotated human judgments about content similarity could foster future research supporting the development of novel similarity measures.

\section*{Acknowledgements}

This work was supported by MTS-Skoltech laboratory on AI.





\bibliographystyle{splncs04}
\bibliography{my.bib}

\begin{thebibliography}{10}
\providecommand{\url}[1]{\texttt{#1}}
\providecommand{\urlprefix}{URL }
\providecommand{\doi}[1]{https://doi.org/#1}

\bibitem{banerjee-lavie-2005-meteor}
Banerjee, S., Lavie, A.: {METEOR}: An automatic metric for {MT} evaluation with
  improved correlation with human judgments. In: Proceedings of the {ACL}
  Workshop. pp. 65--72. Association for Computational Linguistics, Ann Arbor,
  Michigan (2005)

\bibitem{bojanowski-etal-2017-enriching}
Bojanowski, P., Grave, E., Joulin, A., Mikolov, T.: Enriching word vectors with
  subword information. Transactions of the Association for Computational
  Linguistics  \textbf{5},  135--146 (2017)

\bibitem{briakou-etal-2021-evaluating}
Briakou, E., Agrawal, S., Tetreault, J., Carpuat, M.: Evaluating the evaluation
  metrics for style transfer: A case study in multilingual formality transfer.
  In: Proceedings of the 2021 Conference on Empirical Methods in Natural
  Language Processing. pp. 1321--1336. Association for Computational
  Linguistics, Online and Punta Cana, Dominican Republic (2021)

\bibitem{briakou-etal-2021-review}
Briakou, E., Agrawal, S., Zhang, K., Tetreault, J., Carpuat, M.: A review of
  human evaluation for style transfer. In: Proceedings of the 1st Workshop on
  Natural Language Generation, Evaluation, and Metrics (GEM 2021). pp. 58--67.
  Association for Computational Linguistics, Online (2021)

\bibitem{cao-etal-2020-expertise}
Cao, Y., Shui, R., Pan, L., Kan, M.Y., Liu, Z., Chua, T.S.: Expertise style
  transfer: A new task towards better communication between experts and laymen.
  In: Proceedings of the 58th Annual Meeting of the Association for
  Computational Linguistics. pp. 1061--1071. Association for Computational
  Linguistics, Online (2020)

\bibitem{dementieva2021crowdsourcing}
Dementieva, D., Ustyantsev, S., Dale, D., Kozlova, O., Semenov, N., Panchenko,
  A., Logacheva, V.: Crowdsourcing of parallel corpora: the case of style
  transfer for detoxification. In: Proceedings of the 2nd Crowd Science
  Workshop: Trust, Ethics, and Excellence in Crowdsourced Data Management at
  Scale co-located with 47th International Conference on Very Large Data Bases
  (VLDB 2021 (https://vldb.org/2021/)). pp. 35--49. CEUR Workshop Proceedings,
  Copenhagen, Denmark (2021)

\bibitem{devlin-etal-2019-bert}
Devlin, J., Chang, M.W., Lee, K., Toutanova, K.: {BERT}: Pre-training of deep
  bidirectional transformers for language understanding. In: Proceedings of the
  2019 Conference of the North {A}merican Chapter of the Association for
  Computational Linguistics: Human Language Technologies, Volume 1 (Long and
  Short Papers). pp. 4171--4186. Association for Computational Linguistics,
  Minneapolis, Minnesota (2019)

\bibitem{graham-baldwin-2014-testing}
Graham, Y., Baldwin, T.: Testing for significance of increased correlation with
  human judgment. In: Proceedings of the 2014 Conference on Empirical Methods
  in Natural Language Processing ({EMNLP}). pp. 172--176. Association for
  Computational Linguistics, Doha, Qatar (2014)

\bibitem{he2021deberta}
He, P., Liu, X., Gao, J., Chen, W.: Deberta: Decoding-enhanced bert with
  disentangled attention. In: International Conference on Learning
  Representations (2021)

\bibitem{jaccard_index}
Jaccard, P.: Etude de la distribution florale dans une portion des alpes et du
  jura. Bulletin de la Societe Vaudoise des Sciences Naturelles  \textbf{37},
  547--579 (1901)

\bibitem{jhamtani-etal-2017-shakespearizing}
Jhamtani, H., Gangal, V., Hovy, E., Nyberg, E.: Shakespearizing modern language
  using copy-enriched sequence to sequence models. In: Proceedings of the
  Workshop on Stylistic Variation. pp. 10--19. Association for Computational
  Linguistics, Copenhagen, Denmark (2017)

\bibitem{krippendorf}
Krippendorff, K.: Content analysis: An introduction to its methodolog (1980)

\bibitem{krishna-etal-2020-reformulating}
Krishna, K., Wieting, J., Iyyer, M.: Reformulating unsupervised style transfer
  as paraphrase generation. In: Proceedings of the 2020 Conference on Empirical
  Methods in Natural Language Processing (EMNLP). pp. 737--762. Association for
  Computational Linguistics, Online (2020)

\bibitem{lai-etal-2021-thank}
Lai, H., Toral, A., Nissim, M.: Thank you {BART}! rewarding pre-trained models
  improves formality style transfer. In: Proceedings of the 59th Annual Meeting
  of the Association for Computational Linguistics and the 11th International
  Joint Conference on Natural Language Processing (Volume 2: Short Papers). pp.
  484--494. Association for Computational Linguistics, Online (2021)

\bibitem{lee-etal-2021-enhancing}
Lee, D., Tian, Z., Xue, L., Zhang, N.L.: Enhancing content preservation in text
  style transfer using reverse attention and conditional layer normalization.
  In: Proceedings of the 59th Annual Meeting of the Association for
  Computational Linguistics and the 11th International Joint Conference on
  Natural Language Processing (Volume 1: Long Papers). pp. 93--102. Association
  for Computational Linguistics, Online (2021)

\bibitem{lin2004rouge}
Lin, C.Y.: Rouge: A package for automatic evaluation of summaries. In: Text
  summarization branches out. pp. 74--81 (2004)

\bibitem{luo-etal-2019-towards}
Luo, F., Li, P., Yang, P., Zhou, J., Tan, Y., Chang, B., Sui, Z., Sun, X.:
  Towards fine-grained text sentiment transfer. In: Proceedings of the 57th
  Annual Meeting of the Association for Computational Linguistics. pp.
  2013--2022. Association for Computational Linguistics, Florence, Italy (2019)

\bibitem{mikolov2017advances}
Mikolov, T., Grave, E., Bojanowski, P., Puhrsch, C., Joulin, A.: Advances in
  pre-training distributed word representations. arXiv preprint
  arXiv:1712.09405  (2017)

\bibitem{nema-khapra-2018-towards}
Nema, P., Khapra, M.M.: Towards a better metric for evaluating question
  generation systems. In: Proceedings of the 2018 Conference on Empirical
  Methods in Natural Language Processing. pp. 3950--3959. Association for
  Computational Linguistics, Brussels, Belgium (2018)

\bibitem{pang-gimpel-2019-unsupervised}
Pang, R.Y., Gimpel, K.: Unsupervised evaluation metrics and learning criteria
  for non-parallel textual transfer. In: Proceedings of the 3rd Workshop on
  Neural Generation and Translation. pp. 138--147. Association for
  Computational Linguistics, Hong Kong (2019)

\bibitem{papineni2002bleu}
Papineni, K., Roukos, S., Ward, T., Zhu, W.J.: Bleu: a method for automatic
  evaluation of machine translation. In: Proceedings of the 40th annual meeting
  of the Association for Computational Linguistics. pp. 311--318 (2002)

\bibitem{popovic-2015-chrf}
Popovi{\'c}, M.: chr{F}: character n-gram {F}-score for automatic {MT}
  evaluation. In: Proceedings of the Tenth Workshop on Statistical Machine
  Translation. pp. 392--395. Association for Computational Linguistics, Lisbon,
  Portugal (2015)

\bibitem{t5}
Raffel, C., Shazeer, N., Roberts, A., Lee, K., Narang, S., Matena, M., Zhou,
  Y., Li, W., Liu, P.J.: Exploring the limits of transfer learning with a
  unified text-to-text transformer. J. Mach. Learn. Res.  \textbf{21},
  140:1--140:67 (2020)

\bibitem{DBLP:conf/sigir/RaneDLE21}
Rane, C., Dias, G., Lechervy, A., Ekbal, A.: Improving neural text style
  transfer by introducing loss function sequentiality. In: Diaz, F., Shah, C.,
  Suel, T., Castells, P., Jones, R., Sakai, T. (eds.) {SIGIR} '21: The 44th
  International {ACM} {SIGIR} Conference on Research and Development in
  Information Retrieval, Virtual Event, Canada, July 11-15, 2021. pp.
  2197--2201. {ACM} (2021)

\bibitem{rao-tetreault-2018-dear}
Rao, S., Tetreault, J.: Dear sir or madam, may {I} introduce the {GYAFC}
  dataset: Corpus, benchmarks and metrics for formality style transfer. In:
  Proceedings of the 2018 Conference of the North {A}merican Chapter of the
  Association for Computational Linguistics: Human Language Technologies,
  Volume 1 (Long Papers). pp. 129--140. Association for Computational
  Linguistics, New Orleans, Louisiana (2018)

\bibitem{rastogi2020towards}
Rastogi, A., Zang, X., Sunkara, S., Gupta, R., Khaitan, P.: Towards scalable
  multi-domain conversational agents: The schema-guided dialogue dataset. In:
  Proceedings of the AAAI Conference on Artificial Intelligence. vol.~34, pp.
  8689--8696 (2020)

\bibitem{riley-etal-2021-textsettr}
Riley, P., Constant, N., Guo, M., Kumar, G., Uthus, D., Parekh, Z.:
  {T}ext{SETTR}: Few-shot text style extraction and tunable targeted restyling.
  In: Proceedings of the 59th Annual Meeting of the Association for
  Computational Linguistics and the 11th International Joint Conference on
  Natural Language Processing (Volume 1: Long Papers). pp. 3786--3800.
  Association for Computational Linguistics, Online (2021)

\bibitem{sellam-etal-2020-bleurt}
Sellam, T., Das, D., Parikh, A.: {BLEURT}: Learning robust metrics for text
  generation. In: Proceedings of the 58th Annual Meeting of the Association for
  Computational Linguistics. pp. 7881--7892. Association for Computational
  Linguistics, Online (2020)

\bibitem{Yamshchikov_Shibaev_Khlebnikov_Tikhonov_2021}
Yamshchikov, I.P., Shibaev, V., Khlebnikov, N., Tikhonov, A.: Style-transfer
  and paraphrase: Looking for a sensible semantic similarity metric.
  Proceedings of the AAAI Conference on Artificial Intelligence
  \textbf{35}(16),  14213--14220 (2021)

\bibitem{zhang2019bertscore}
Zhang, T., Kishore, V., Wu, F., Weinberger, K.Q., Artzi, Y.: Bertscore:
  Evaluating text generation with bert. arXiv preprint arXiv:1904.09675  (2019)

\bibitem{zhuang-etal-2021-robustly}
Zhuang, L., Wayne, L., Ya, S., Jun, Z.: A robustly optimized {BERT}
  pre-training approach with post-training. In: Proceedings of the 20th Chinese
  National Conference on Computational Linguistics. pp. 1218--1227. Chinese
  Information Processing Society of China, Huhhot, China (2021)

\end{thebibliography}

\end{document}